\documentclass[10pt,letterpaper]{article}

\usepackage[top=5cm, bottom=5cm, left=3.8cm, right=3.8cm]{geometry} 

\usepackage{mathptmx}
\usepackage{hyperref}
\usepackage{url}
\usepackage{subcaption}
\usepackage[breakable, theorems, skins]{tcolorbox}

\usepackage{courier}

\usepackage{indentfirst}

\usepackage{booktabs}

\usepackage{graphicx}

\usepackage[font=small,labelfont=bf]{caption}

\usepackage{titling}
\pretitle{\begin{center}\fontsize{16pt}{19pt}\selectfont\bfseries}
\posttitle{\par\end{center}\vskip 1ex}

\renewenvironment{abstract}
{\begin{quote}
 \noindent {\small \bfseries \abstractname.} \small
}
{\medskip\noindent
 \end{quote}
}

\newenvironment{keywords}
{\begin{quote}
 \noindent {\small \bfseries Keywords: } \small
}
{\medskip\noindent 
 \end{quote}
}

\usepackage{titlesec}

\titleformat*{\section}{\large\bfseries}
\titleformat*{\subsection}{\normalsize\bfseries}
\titleformat*{\subsubsection}{\normalsize\bfseries}
\titleformat*{\paragraph}{\normalsize\bfseries}
\titleformat*{\subparagraph}{\normalsize\bfseries}

\usepackage[plain, noline]{algorithm2e}
\SetAlFnt{\fontfamily{pcr}\selectfont} 
\SetAlCapSkip{2ex}
\IncMargin{-1em}

\SetKwInput{KwIn}{Input}
\SetKwInput{KwOut}{Output}
\SetKwIF{If}{ElseIf}{Else}{if}{then}{else\ if}{else}{end\ if}
\SetKwFor{For}{for}{do}{end\ for}
\SetKw{KwTo}{until}
\SetKwRepeat{Repeat}{repeat}{until}
\SetKwFor{While}{while}{do}{end\ while}

\usepackage[hang,flushmargin]{footmisc} 

\usepackage{bibspacing}
\begin{document}

\title{Sustainable Artificial Intelligence through Continual Learning} 

\author{
\normalsize Andrea Cossu$^1$, Marta Ziosi$^2$, Vincenzo Lomonaco$^3$ \\ 
\small $\lbrace$andrea.cossu@sns.it$^1$, marta.ziosi@aiforpeople.org$^2$, vincenzo.lomonaco,@aiforpeople.org$^3$ $\rbrace$ 
\and 
\small AI for People$^{2, 3}$, University of Pisa$^{1,3}$, University of Oxford$^2$, Scuola Normale Superiore$^1$
}

\date{}

\pagestyle{empty}

\maketitle
\thispagestyle{empty}

\vspace{-3.75em}

\begin{abstract}
The increasing attention on Artificial Intelligence (AI) regulation has led to the definition of a set of ethical principles grouped into the Sustainable AI framework. In this article, we identify Continual Learning, an active area of AI research, as a promising approach towards the design of systems compliant with the Sustainable AI principles. While Sustainable AI outlines general desiderata for ethical applications, Continual Learning provides means to put such desiderata into practice.
\end{abstract}


\section{Introduction}

\begin{figure}[t] 
\hspace{20px}
\begin{subfigure}[t]{0.35\textwidth}
  \centering
  \includegraphics[width=\textwidth]{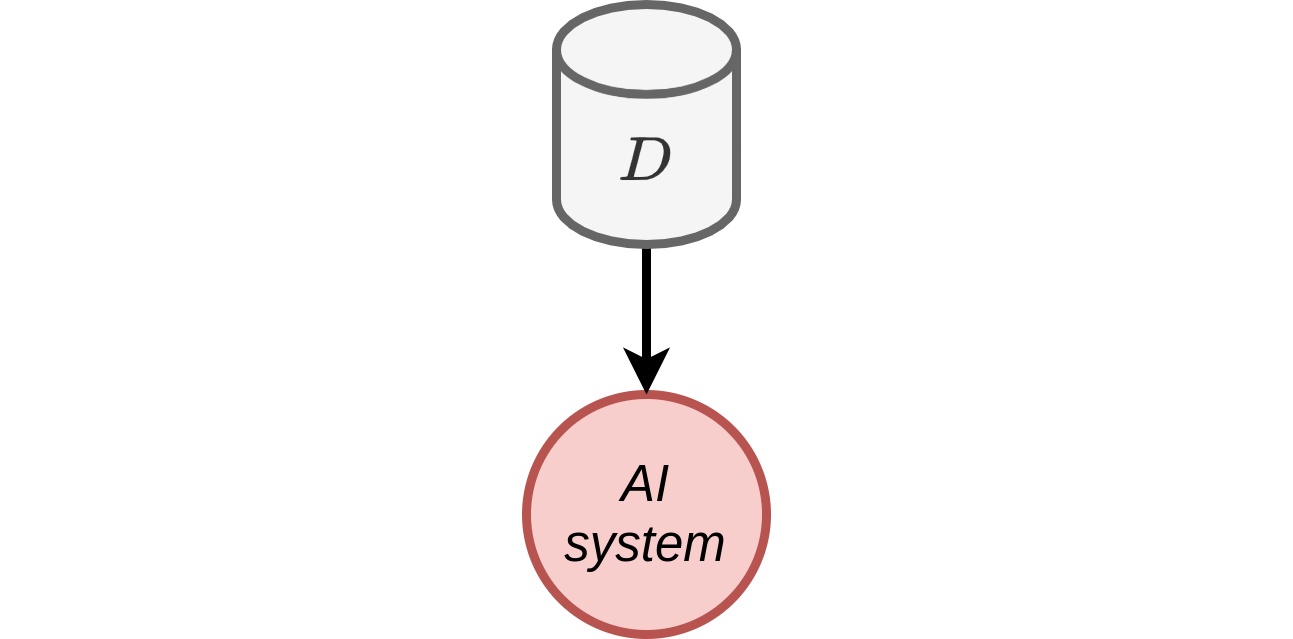}
  \caption{Offline AI system}
  \label{fig:ai}
  \end{subfigure}
  %\hfill
  \hspace{60px}
  \begin{subfigure}[t]{0.35\textwidth}
  \centering
  \includegraphics[width=\textwidth]{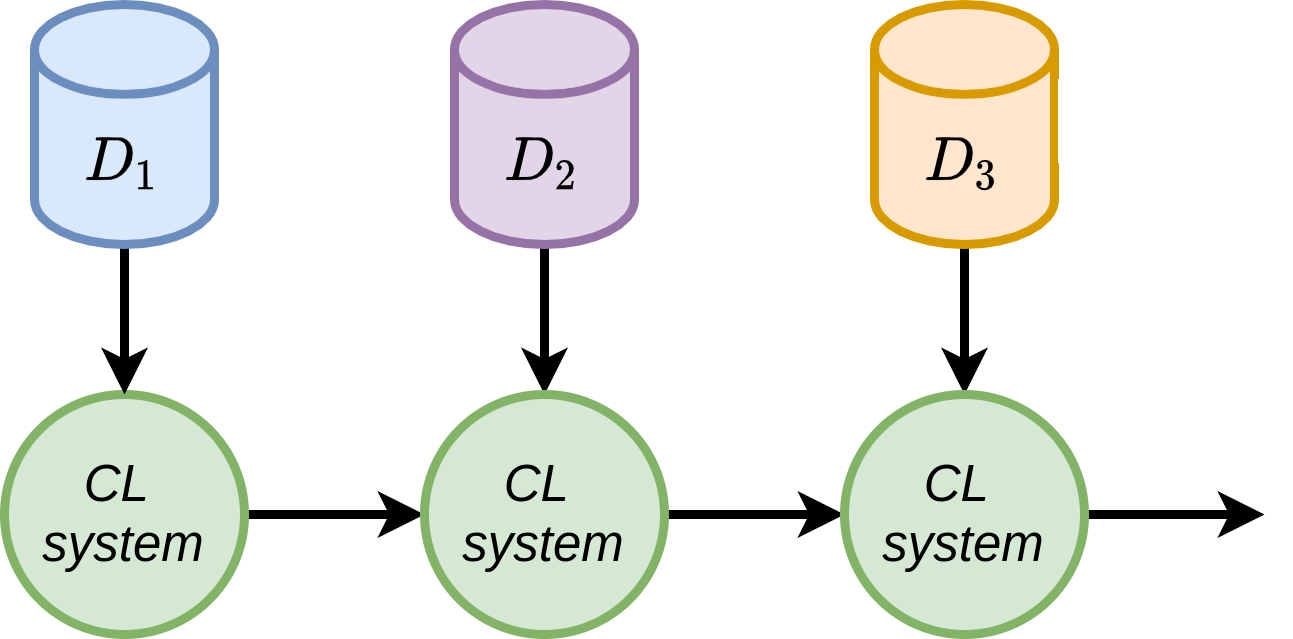}

  \caption{Continual learning system}
  \label{fig:cl}
  \end{subfigure}
  \caption{Difference between offline AI (left) and Continual Learning setup (right). While offline learning learns in a single phase from a static set of data, Continual Learning systems learn from a stream of non-stationary data.}
  \label{fig:setup}
\end{figure}

AI systems learning from raw data are widely used in a large number of real world applications \cite{lecun2015}. Their performance has dramatically increased the degree to which we are able to address previously inaccessible tasks (e.g. machine translation, image segmentation, speech-to-text and many others). Fortunately, the debate around AI is not only centered on its performance, but it is also increasingly focusing on other \textit{non-functional} properties, like explainability, privacy and scalability \cite{jobin2019}. In particular, the Sustainable AI framework encompasses a set of principles which are often deemed too general and abstract to lead to concrete implementation means \cite{vanwynsberghe2021}. 
In this paper, we focus on the "Sustainability of AI" and we show how Continual Learning \cite{lesort2020}, a recently growing topic in AI, can be a suitable candidate to render current AI Ethics principles both more operationalisable (one of their main critiques), and more socially, financially and environmentally sustainable. 
We proceed by first reviewing the Sustainable AI framework and then by presenting the main characteristics of the Continual Learning objective. Finally, we show the tight correspondence between Sustainable AI and Continual Learning objectives and we discuss how future research on the latter could lead to systems able to respect the Sustainable AI principles previously described.

\section{Sustainable AI Principles} \label{sec:sustainable}
As the range of AI capabilities expands, so does our awareness of the ethical issues related to the the design, development, deployment and
use of AI systems. The promise for positive change that AI represents has been challenged by several reports on ethically questionable uses of AI in contexts as varied as healthcare, education, law enforcement, recruitment, risk-assessment and more. This increasing number of reports has contributed to the birth of the field of AI Ethics. AI Ethics refers to "the study of ethical and societal issues facing developers, producers, consumers, citizens, policy makers, and civil society organizations" in the ambit of AI systems \cite{vanwynsberghe2021}. The field has so far taken a principled-base approach to ethics, with more than 70 documents containing AI Ethics principles published between 2016 and 2019. This principles stand as abstractions, arising as responses to practical concerns around the application of AI techniques \cite{turilli2007}. They are expressed by values such as fairness, privacy, accountability, transparency, etc. One of the main critiques advanced in relation to these AI Ethics principles has been their overtly abstract character, represented by the failure to move from principles to practice \cite{morley2020}. This critique led to a search for more usable solutions.

Concurrently, Wynsberghe in \cite{vanwynsberghe2021} has started calling for a change of focus in AI Ethics. She calls for a focus on "Sustainable AI" and she conceives it as a new wave of AI Ethics that puts sustainable development at its score. Sustainable AI encompasses both AI for Sustainability, represented by initiatives such as "AI for Good"  or "AI4SDGs", as well as the Sustainability of AI represented by underfunded concerns about the social, financial and environmental impact of AI systems. While the initiatives on AI for good abound, AI Ethics has paid much less attention to assessing what Wynsberghe calls the "Sustainability of AI", especially in terms of the environmental impact of AI systems themselves. Here, we present the set of AI Ethics principles that we will be concerned with in this paper:

\begin{itemize}
    \item Efficiency \& Scalability. Modern machine learning solutions usually require long training phase on large datasets. Training such models have a drastic impact on both the environment (due to high resource consumption) and the dynamic acquisition of new information (due to the large amount of data required). Sustainable AI approaches, instead, should incorporate an efficient and scalable approach to learn ever complex information from a stream of data, without requiring massive amount of computation each time new knowledge comes in.
    \item Fairness, Privacy \& Security.
    While these three principles can stand in separation, they relate to each other at some level. Fairness can be understood as the absence of bias and discrimination, that is, ensuring equitability \cite{jobin2019}. This may refer to the inner process of algorithmic decisions as well as in their outcomes. Responses to discrimination, as well as to biased and opaque decision-making often appeal to data privacy \cite{mittelstadt2016}. This can entail solutions that prevent decisions being taken based on protected attributes such as race or gender. Privacy refers to the capacity for an individual to control information about themselves. AI Ethics guidelines present privacy both as a value to uphold and as a right to be protected \cite{jobin2019}. One way in which privacy can be violated is by breaching the security of a system through adversarial attacks that expose the model structure, the training or test dataset. Security here refers to measures taken to avoid foreseeable as well as unintentional harm from or through AI systems. 
    \item Accuracy \& Robustness.
    Algorithmic conclusions are based on probabilities. As such, they are not infallible and they might incur in errors during execution. The fact that algorithmic decisions may be taken on inconclusive evidence renders the accuracy of their predictions an important value to uphold. Apart from accurate, AI systems need to be resilient and secure. This aspects are expressed in the principle of robustness.
    \item Explainability, Transparency \& Accountability.
    Explainability refers to the explanation of an algorithmic decision being comprehensible to technical people such as data-scientists as well as to non-technical people such as the data subjects affected by the decision \cite{mittelstadt2016}. While ethical guidelines feature an overreliance on explicability \cite{morley2020}, it can be problematic to uphold this value without information being accessible. Accessibility and comprehensibility of information are the two main components of transparency \cite{turilli2009}. Transparency stands in contrast to the opacity of algorithms, especially when referring to the problem of black-box algorithms. Additionally, the opacity of machine learning algorithms is considered as a challenge to accountability \cite{schermer2011}. In that respect, solutions to ensure accountability such as human oversight are threatened by opaque algorithms. There is no accountability without transparency.
\end{itemize}

\section{Continual Learning as Key Technological Enabler} \label{sec:cl}

Making machines that can learn and think like humans has always been one of the main goal of Artificial Intelligence (AI) research. Since its inception, AI has been thought as grounded in the ability to learn and adapt to new situations. Indeed, learning itself progresses through a continuous exposure to external stimuli and experiences. However, despite early speculations and few pioneering works, very little research and effort has been devoted to address this vision. Current AI systems (Figure \ref{fig:ai}) learn offline from a static dataset and greatly suffer from the exposure to new data or environments which even slightly differ from the ones for which they have been trained for \cite{french1999, grossberg1980}. One notable example of such deterioration is the catastrophic forgetting phenomenon, which causes the disruption of previous knowledge when learning new information \cite{french1999}. Moreover, the learning process is usually constrained on fixed datasets within narrow and isolated tasks which may hardly lead to the emergence of more complex and autonomous intelligent behaviors. 
The idea of creating algorithms that can learn continually and incrementally from a progressive exposure to external stimuli (Figure \ref{fig:cl}) is at the core of the recent surging of interest in Continual Learning research, within the AI and Machine Learning community \cite{maltoni2019}. 
Learning continually means not only being able to adapt to unpredictable circumstances not foreseen during the design of our AI systems, but also to build systems that are \emph{efficient}, \emph{scalable} and \emph{easily amendable}, or in other words, \emph{sustainable}.

\subsection{Continual Learning for Sustainable AI}
We will now address the sustainable AI principles described in Section \ref{sec:sustainable} and we will show how they can be satisfied by means of AI systems which learn continuously over time.

\paragraph{Efficiency and Scalability} The efficiency of AI systems is usually measured with respect to metrics like \textit{computing time} (how much time does it take to learn), \textit{memory occupancy} (how much memory is it required to learn) and \textit{sample efficiency} (how many data patterns are required to learn) \cite{lesort2020, lomonaco2021}. Scalability is commonly approached by extrapolating these metrics for growing dataset / model sizes. However, the evaluation protocol currently in use for both efficiency and scalability considers only static environments, where the training process occurs on a fixed dataset, entirely available beforehand. Continual learning operates in a completely different settings, where patterns are learned \emph{over time}: the computing time and memory occupancy are distributed during the entire learning experience and, since the model is capable to tackle few patterns at a time incrementally, available resources can be adjusted dynamically, as needed. The advantage in efficiency and scalability for Continual Learning with respect to offline AI models increases with the nonstationary degree of the environment: each new piece of information may require to retrain a non continual model on both new and previous data. This may be unnecessary and wasteful, since most of the knowledge needed is already inside the model. Continual learning, on the contrary, only requires to train a model on the new incoming samples, possibly exploiting temporal coherence between successive patterns to improve performance \cite{parisi2020, cossu2021}.

\paragraph{Fairness, Privacy \& Security} AI systems operating on the edge (e.g., on device) are not able to incorporate new information, since they suffer from the catastrophic forgetting of previous knowledge. The process of retraining on both new and old data is expensive due to the large size of datasets and cannot be performed on resources-constrained devices like smartphones or even modern laptop. Therefore, each time new data is available the model need to be sent back to a central computing unit (e.g., datacenter on cloud) on which retraining is performed. This process exposes security holes which can be exploited by an external attacker in order to retrieve sensitive knowledge. For example, a wearable device collecting medical parameters from a patient may need to send its data on the cloud for further training, thus allowing for stealing of private information. Continual learning systems are instead able to learn directly on-device and they remove the need to send data to external infrastructures, greatly limiting possible security issues. \\
On a different note, lack of fairness in model predictions often comes by the fact that AI models are trained on biased datasets, where information is incomplete or driven by social/political prejudices. Continual Learning may help in understanding and preventing biased predictions in two ways: firstly, it allows to study changes in predictions as new data is injected into the model. Since no retraining is performed, the change in prediction is directly related to the input patterns provided to the model. Secondly, once biased prediction is detected, Continual Learning is capable of efficiently incorporate new information. In real-world applications, this opportunity may be prevented by the fact that the model would need a long and costly update, possibly to manage few edge cases. 

\paragraph{Accuracy \& Robustness} While current AI systems excel in addressing narrow tasks with high accuracy, they are prone to catastrophic forgetting of previous knowledge as soon as they encounter new information. Therefore, the prediction accuracy with respect to all the data seen by the model reduces drastically when considering even slightly nonstationary environments. This makes AI systems unreliable and not robust to possible changes in the incoming data. In order to address this point, Continual Learning studies the tradeoff between the prediction accuracy on the current data and the prediction accuracy on all previous data. Even if existing Continual Learning solutions do not completely remove catastrophic forgetting of previous knowledge, they are able to largely mitigate it \cite{delange2021}. As a matter of fact, catastrophic forgetting may never be solved completely. It is nonetheless fundamental to highlight that accuracy alone is not a robust evaluation metric and that the performance should be measured in terms of the entire model lifetime. This directly advocates for the use of effective continual learning models able to operate in less confined environments. 

\paragraph{Explainability, Transparency \& Accountability}
Modern AI solutions are often thought as \emph{black boxes}, with few to no possibility to understand the reasons and the rationale behind their decisions. Continual Learning does not directly address this problem, since a continual learner may as well be able to update its internal representations during time without disclosing any hints on what such representations mean. Interestingly, a prominent line of research in Continual Learning calls for the use of neuroscience-grounded solution \cite{cui2016}. While the human brain is far from being explainable and accountable, it is nonetheless possible in that context to use scientific tools to draw general conclusions on the behavior of different brain regions and to study which of them are more entrusted with a high-level cognitive function. By building Continual Learning models which follow neuroscience principles, it may be possible in the future to study these systems with similar tools and to reach a similar understanding of its inner functioning, which seems one of the most ambitious advances towards explainable systems.

\section{Conclusion and Future Directions}
In this paper, we presented Continual Learning as an effective approach towards the development of Sustainable AI systems able to tackle real-world applications in respect to a set of principles including efficiency, privacy, robustness and explainability. In particular, we showed that the Continual Learning objectives and the Sustainable AI desiderata are aligned, in the sense that a model able to learn continuously will be most likely brought to respect the Sustainable AI principles. The growing amount of contributions to the field suggests that the path is worth pursuing: as new Continual Learning solutions are designed, the advantages in the use of models which learn continuously over static ones will become wider and will highlight even further the opportunities for a concrete realization of the Sustainable AI framework. While we acknowledge that Continual Learning may not be the only viable solution towards sustainable models, here we supported the idea that Continual Learning is currently the most promising candidate and a necessary requirement to drive research in the direction of Sustainable AI.

%\newpage

% The next two lines define the bibliography style to be used, and the bibliography file.
\bibliographystyle{vancouver}
\bibliography{CL, ethics}

\end{document}